\documentclass{article}

\usepackage{xcolor}
\usepackage{comment}
\usepackage{microtype}
\usepackage{graphicx}
\usepackage{subcaption}
\usepackage{booktabs} 

\usepackage{hyperref}

\usepackage[preprint]{icml2026}

\usepackage{amsmath}
\usepackage{amssymb}
\usepackage{mathtools}
\usepackage{amsthm}
\usepackage{pifont}
\newcommand{\cmark}{\ding{51}}
\newcommand{\xmark}{\ding{55}}

\usepackage[capitalize,noabbrev]{cleveref}

\theoremstyle{plain}

\theoremstyle{definition}

\theoremstyle{remark}

\usepackage[textsize=tiny]{todonotes}

\icmltitlerunning{CD-LoRA: Consistency-Driven Low-Rank Adaptation for Multi-Task Fine-Tuning}

\begin{document}

\twocolumn[
  \icmltitle{CD-LoRA: Consistency-Driven Low-Rank Adaptation for Multi-Task Fine-Tuning}



  \icmlsetsymbol{equal}{*}

  \begin{icmlauthorlist}
    \icmlauthor{Qian Zha}{yyy}
    \icmlauthor{Jinda Liu}{yyy}
    \icmlauthor{Yuan Wu}{yyy}
    \icmlauthor{Yi Chang}{yyy,comp,sch}
  \end{icmlauthorlist}

  \icmlaffiliation{yyy}{School of Artificial Intelligence, Jilin University}
  \icmlaffiliation{comp}{Engineering Research Center of Knowledge-Driven Human-Machine Intelligence, MOE, China}
  \icmlaffiliation{sch}{International Center of Future Science, Jilin University}

  \icmlcorrespondingauthor{Yuan Wu}{yuanwu@jlu.edu.cn}

  \icmlkeywords{Machine Learning, ICML}

  \vskip 0.3in
]



\printAffiliationsAndNotice{}  

\begin{abstract}
While Multi-Task Learning (MTL) is essential for adapting Large Language Models (LLMs) to diverse domains,
prevailing LoRA-based methods rely on complex routing mechanisms that partition task-specific knowledge.
In this work, we reveal that such routing-based designs are \textbf{prone to a training–inference discrepancy},
where stochastic routing decisions under distribution shifts compromise inference stability.
Driven by a second-order Taylor analysis that exposes the instability induced by routing variance,
we challenge the training-inference discrepancy and propose \textbf{Consistency-Driven Low-Rank Adaptation (CD-LoRA)}.
By eliminating routers entirely, CD-LoRA employs a consistency-driven alignment mechanism to enforce
\textbf{representation congruence} across tasks in a shared low-rank space.
This paradigm fosters robust, task-agnostic features without explicit partitioning overhead.
Extensive experiments show that CD-LoRA consistently outperforms state-of-the-art multi-adapter baselines,
offering a simpler, router-free, and more stable solution for multi-task PEFT.
The code is available at the anonymous link \href{https://github.com/zhaqian21/CD-LoRA}{CD-LoRA}.
\end{abstract}

\section{Introduction}
LLMs have demonstrated remarkable performance across a wide range of
Natural Language Processing (NLP) tasks \citep{llm, llm_survey, llm_evaluation_survey}.
Despite their strong zero-shot generalization, adapting LLMs to specialized domains or evolving tasks
remains essential \citep{BioBERT, training, Bloomberggpt}.
While Supervised Fine-Tuning (SFT) is the standard adaptation paradigm,
Full Parameter Fine-Tuning (FPT) incurs prohibitive computational and memory costs as model scales grow
\citep{parameter-efficient, Parameter-efficient_fine-tuning}.
This has motivated extensive research on Parameter-Efficient Fine-Tuning (PEFT)\citep{lora, prefix_tuning, adapter, parameter-efficient},
among which Low-Rank Adaptation (LoRA)\citep{lora} has emerged as a dominant approach.

\begin{figure*}[t]
    \centering
    \includegraphics[width=1.0\linewidth]{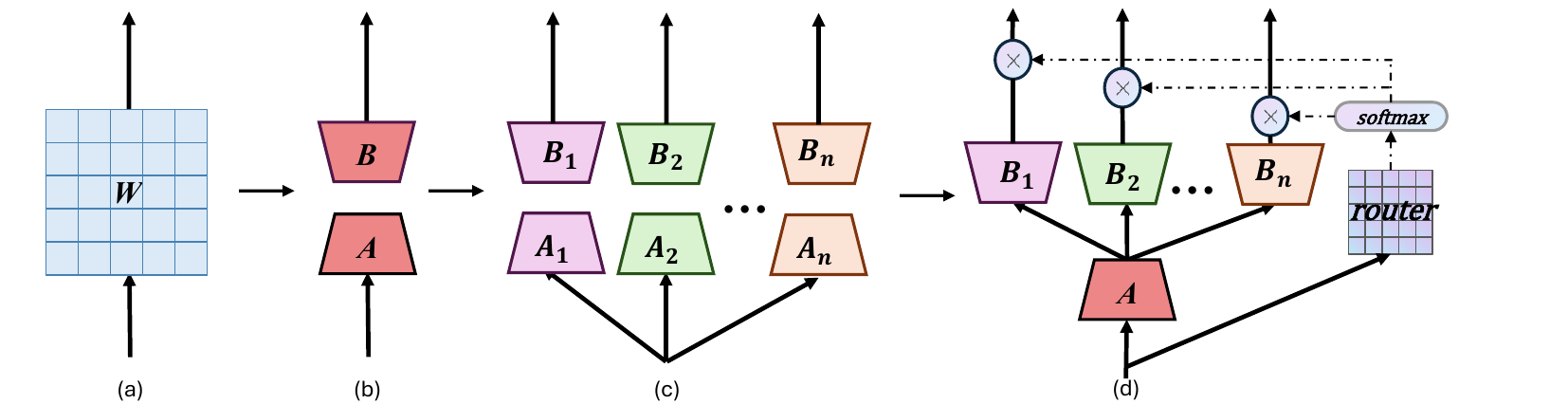}
    \caption{Architectural comparison of fine-tuning paradigms.
    (a) Full Parameter Fine-Tuning (FPT);
    (b) Vanilla LoRA;
    (c) Multi-Adapter architectures with task-specific low-rank modules;
    (d) Multi-Head or asymmetric LoRA variants with routing mechanisms.}
    \label{fig:arch_comparison}
\end{figure*}

Recent efforts~\citep{Hydralora,R-LoRA} extend LoRA to Multi-Task Learning (MTL) by introducing task-specific or multi-head
adapter structures.
Early multi-adapter methods allocate independent low-rank parameters per task \citep{OrchMoE},
while subsequent designs improve parameter efficiency by sharing components across tasks, such as a common $A$ with multiple $B_{i}$s \citep{Hydralora}.
To further enhance flexibility, many approaches incorporate routing mechanisms inspired by
Mixture-of-Experts (MoE)~\citep{MoE}, dynamically weighting adapter outputs and encouraging head diversity
\citep{Mixlora, LoraHub, MoLA, R-LoRA}. Figure~\ref{fig:arch_comparison} summarizes the architectural evolution of representative
fine-tuning paradigms.

Collectively, these methods assume that effective multi-task adaptation relies on \textbf{dynamic routing}(as shown in subfigure (d) of Figure~\ref{fig:arch_comparison}.) to disentangle \textbf{task-specific knowledge}.

However, our work challenges this paradigm. We analytically identify a \textit{training--inference discrepancy} (Section~\ref{sec:theoretical})---an inherent divergence between training and inference objectives exacerbated by routing non-linearity. Beyond theoretical derivation, latent space visualizations (Section~\ref{sec:Analysis}) confirm that this discrepancy induces significant performance degradation. 

Motivated by this analysis, we propose \textbf{Consistency-Driven Low-Rank Adaptation (CD-LoRA)}, which effectively mitigates the training--inference discrepancy in multi-task PEFT. 
Extensive experiments show that CD-LoRA not only consistently outperforms multi-adapter baselines but also moderately reduces computational overhead.

Our contributions are summarized as follows:
\begin{itemize}
    \item \textbf{Formalization of Training--Inference Discrepancy:} We analytically unveil that non-linear routing in multi-head LoRA architectures inherently induces a \textit{training--inference discrepancy}, which undermines model stability\footnote{In this context, \textit{stability} refers to the consistency of training and inference.} under stochastic decision-making.
    \item \textbf{Empirical Re-evaluation of Routing Mechanisms:} We challenge the conventional reliance on expert partitioning by demonstrating that a streamlined, router-free paradigm not only mitigates this discrepancy but consistently outperforms complex routing-based baselines.
    \item \textbf{A Consistency-Driven Perspective for MTL:} We shift the multi-task adaptation focus from structural isolation to representation congruence, identifying that maintaining consistency in a shared low-rank space is more critical for performance than explicit task-specific partitioning.
    \item \textbf{The CD-LoRA Framework:} We propose CD-LoRA, a plug-and-play, router-free PEFT framework. Specifically, CD-LoRA eliminates architectural overhead while achieving state-of-the-art performance and enhanced robustness across diverse benchmarks.
\end{itemize}

\section{Related Work}
\subsection{Low-Rank Adaptation (LoRA)}
\label{sec:lora}


LoRA \citep{lora} is designed to efficiently adapt LLMs by approximating the weight updates $\Delta \mathbf{W}$ of dense matrices within Transformer blocks \citep{Attention}. Built on the hypothesis that such weight updates exhibit low intrinsic dimensionality, LoRA decomposes $\Delta \mathbf{W}$ into two low-rank matrices, namely LoRA-$\mathbf{A} \in \mathbb{R}^{r \times n}$ and LoRA-$\mathbf{B} \in \mathbb{R}^{m \times r}$, where $\mathbf{W} \in \mathbb{R}^{m \times n}$ denotes the original frozen weight matrix of the model. The rank $r$ is deliberately set to be much smaller than $\min(m, n)$, which reduces the number of trainable parameters from $\mathcal{O}(mn)$ (the order of magnitude for full dense matrix updates) to $\mathcal{O}(r(m+n))$, achieving significant parameter efficiency.
For a given input $\mathbf{x}$, the forward propagation process is adjusted as shown:

\begin{equation}
    \mathbf{h} = (\mathbf{W} + \Delta \mathbf{W})\mathbf{x} = \mathbf{W}\mathbf{x} + \mathbf{B}\mathbf{A}\mathbf{x},
    \label{eq:lora_forward}
\end{equation}

Here, $\Delta \mathbf{W} = \mathbf{B}\mathbf{A}$ represents the low-rank weight update induced by LoRA. A pivotal advantage of this approach is that post-training, the low-rank update $\Delta \mathbf{W}$ can be seamlessly merged back into the original weight matrix $\mathbf{W}$, resulting in zero additional inference overhead compared to the base model without adaptation.

Several extensions have been proposed on top of the original LoRA framework.
AdaLoRA~\citep{Adalora} dynamically allocates rank budgets across layers,
while DoRA~\citep{Dora} decomposes weight updates into magnitude and
direction components. Align-LoRA~\cite{align} align representation distributions across tasks to learn task-shared knowledge and boost multi-task generalization, theoretically tightening the generalization bound by minimizing inter-task distribution discrepancies. Other methods, such as PiSSA~\citep{Pissa} and NLoRA~\citep{NLoRA},
focus on improving performance through better initialization strategies and
alternative decomposition schemes, highlighting ongoing efforts to enhance the
effectiveness of LoRA-based adaptation.

\subsection{LoRA for Multi-Task Learning.}
\begin{table*}
\centering
\caption{Architectural Variants of LoRA for Multi-Task Learning}
\label{tab:lora_architectures}
\begin{tabular}{lccc}
\toprule
\textbf{Architecture} 
& \textbf{Multi-Adapter} 
& \textbf{Multi-Head} 
& \textbf{Dynamic Routing} \\
\midrule
Weight Update 
&
$\displaystyle \Delta W = \sum_{i=1}^{N} B_i A_i$
&
$\displaystyle \Delta W = \sum_{i=1}^{N} B_i A$
&
$\displaystyle \Delta W = \sum_{i=1}^{N} \omega_i(x)\, B_i A$ \\
\bottomrule
\end{tabular}
{\footnotesize\\
\textit{\textbf{Notation:}  $ \omega_i(x) $  denotes input-dependent mixing weights, typically produced by a softmax-based gating network.}}
\label{tab:Architectural}
\end{table*}

Extending LoRA to multi-task learning naturally motivates the introduction of multiple trainable components to accommodate task heterogeneity.
Early approaches adopt a \textbf{Multi-Adapter} paradigm \citep{Mixlora, MoLA}, where independent LoRA adapters, specifically separate $(B_i, A_i)$ matrix pairs, are allocated for each task. Though effective, this design causes the number of parameters to grow linearly with the number of tasks, and the strict parameter isolation hinders cross-task knowledge transfer \cite{LoraHub}.

To improve parameter efficiency, subsequent studies~\citep{Hydralora,R-LoRA} exploit the asymmetric roles of LoRA’s low-rank factors.
Empirical evidence suggests that the down-projection  LoRA-A primarily captures task-agnostic and redundant representations, whereas the up-projection LoRA-B encode task-specific variations~\citep{R-LoRA}.
This insight gives rise to the \textbf{Multi-Head} architecture, which shares a single projection  LoRA-A across tasks while maintaining multiple task-specific heads $\{B_i\}$.
Compared to multi-adapter designs, multi-head LoRA substantially reduces parameter overhead while enabling partial knowledge sharing.

To further enhance task specialization within the multi-head paradigm, several methods~\cite{Mixlora, LoraHub, MoLA, R-LoRA} introduce \textbf{Dynamic Routing} mechanisms inspired by the MoE framework.

In Table~\ref{tab:Architectural}, we summarize the weight update formulations of different LoRA architectures for multi-task learning.

Such routing mechanisms adaptively select or combine LoRA heads based on the input, aiming to disentangle task-specific knowledge and improve flexibility.

However, this added expressiveness comes at a cost.
Input-dependent routing prevents the merged update $\Delta W$ from being pre-computed and folded into the frozen backbone after training.
As a result, both the routing network and multiple LoRA heads must be evaluated at inference time, introducing non-negligible latency and memory overhead.
This trade-off undermines one of LoRA’s key practical advantages \textit{zero inference overhead} and raises concerns about \textbf{training–inference consistency} in dynamically routed architectures.

\section{Motivation}
\label{sec:motivation}

\subsection{Formalizing the Training--Inference Discrepancy}
\label{sec:theoretical}

To examine whether complex routing mechanisms are indeed necessary for multi-task PEFT, we begin by formalizing the mismatch between
training and deployment, referred to as the
\emph{training--inference discrepancy}.

In routing-based LoRA variants such as R-LoRA~\citep{R-LoRA}, the training-time output
for an input $\mathbf{x} \in \mathbb{R}^{n}$ under a stochastic dropout mask
$\mathbf{z} \sim p(\mathbf{z})$ is given by
\begin{equation}
    \Delta W(\mathbf{x}, \mathbf{z}) = \sum_{i=1}^{N} \omega_{i}(\mathbf{x}, \mathbf{z}) \cdot
    \mathbf{B}_{i} \mathbf{A}(\mathbf{x} \odot \mathbf{z}),
    \label{eq:routing_output}
\end{equation}
where $N$ denotes the number of LoRA heads,
$\mathbf{A} \in \mathbb{R}^{r \times n}$ is the shared down-projection matrix,
$\mathbf{B}_{i} \in \mathbb{R}^{m \times r}$ is the up-projection matrix of the $i$-th head,
and $\omega_i(\cdot)$ is a \textbf{non-linear} routing weight, typically produced by a \textbf{softmax-based
gating network}. The operator $\odot$ denotes element-wise multiplication.

At inference time, a deterministic mask
$\bar{\mathbf{z}} = \mathbb{E}[\mathbf{z}]$ is used.
Due to the \textbf{non-linearity} of $\omega_i(\cdot)$ which includes a \textbf{softmax operation}, the expected training output does not
generally coincide with the inference-time output, i.e.,
$\mathbb{E}_{\mathbf{z}}[f(\mathbf{x}, \mathbf{z})] \neq f(\mathbf{x}, \bar{\mathbf{z}})$.
This mismatch gives rise to a training--inference discrepancy that may degrade
predictive stability.

\paragraph{Discrepancy Decomposition in Router-Free Architectures.}
To isolate the contribution of routing, we consider a router-free multi-head LoRA
variant in which all heads are uniformly weighted.
The resulting training--inference discrepancy for input $\mathbf{x}$ is defined as
\begin{equation}
    \mathrm{TID}(\mathbf{x}) =
    \left\|
    \frac{1}{N} \sum_{i=1}^{N} \mathbf{B}_{i}
    \left(
    \mathbb{E}_{\mathbf{z}}[\mathbf{A}(\mathbf{x} \odot \mathbf{z})]
    - \mathbf{A}(\mathbf{x} \odot \bar{\mathbf{z}})
    \right)
    \right\|,
    \label{eq:training_inference}
\end{equation}
where $\|\cdot\|$ denotes either the Frobenius norm.

Applying the sub-multiplicativity of matrix norms yields the following upper bound:
\begin{equation}
    \mathrm{TID}(\mathbf{x}) \le
    \left( \frac{1}{N} \sum_{i=1}^{N} \|\mathbf{B}_{i}\| \right)
    \cdot
    \left\|
    \mathbb{E}_{\mathbf{z}}[\mathbf{A}(\mathbf{x} \odot \mathbf{z})]
    - \mathbf{A}(\mathbf{x} \odot \bar{\mathbf{z}})
    \right\|.
    \label{eq:gap_bound}
\end{equation}

This decomposition shows that, once routing is removed, the training--inference
discrepancy is governed by two factors: the aggregate scale of the up-projection
matrices $\{\mathbf{B}_i\}_{i=1}^{N}$ and the sensitivity of the shared down-projection
matrix $\mathbf{A}$ to stochastic perturbations.
Notably, the discrepancy is entirely attributed to instability in the low-rank
projection stage, rather than head selection or routing decisions.

\subsection{Empirical Analysis of Representation Structure}
\label{sec:Analysis}
To empirically examine the representation behavior suggested by our theoretical
analysis, we study the structure of activations produced by the LoRA-A
(\texttt{down\_proj}) module under 12-task joint training.
The \texttt{down\_proj} module corresponds to the shared low-rank projection matrix
$\mathbf{A}$, which maps high-dimensional hidden states into a compact intrinsic
subspace before task-specific transformations.
As shown in Section~\ref{sec:theoretical}, instability in this shared projection is
the primary source of the training--inference discrepancy, making it a natural
candidate for representation-level analysis.

We extract layer-wise LoRA-A activations from the \texttt{down\_proj} module and
visualize their distributions using t-SNE.
This analysis focuses on whether representations induced by different tasks remain
separable or collapse into a shared manifold when projected into the low-rank space.

\begin{figure}[ht]
    \centering
    \includegraphics[width=0.8\linewidth]{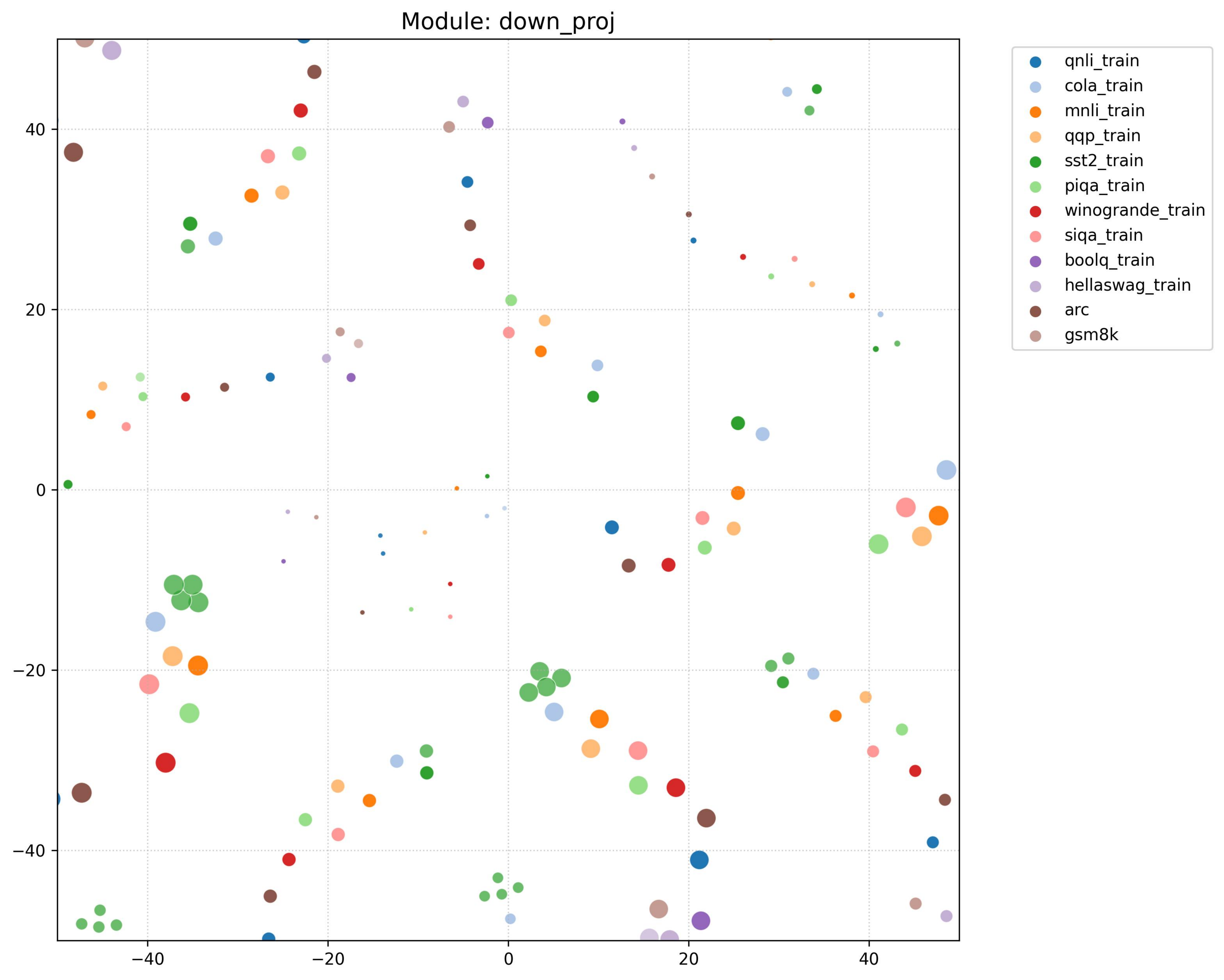}
    \caption{t-SNE visualization of LoRA-A activations in the \texttt{down\_proj} module.
    Different colors indicate different tasks.
    Representations from multiple tasks exhibit substantial overlap in the shared
    low-rank space.}
    \label{fig:tsne_comparison}
\end{figure}

As illustrated in Figure~\ref{fig:tsne_comparison}, LoRA-A activations from different
tasks are highly intermingled, forming a largely overlapping structure rather than
well-separated clusters.
Although t-SNE provides only a qualitative perspective, this observation aligns with
our theoretical findings: the limitation of multi-task adaptation does not primarily
stem from insufficient task partitioning, but from instability within the shared
projection space.
Detailed visualization settings and extraction procedures are provided in Appendix~\ref{app:tsne_setup}.
\section{Method}
\subsection{From Dynamic Routing to Unified Low-Rank Representations}
\label{sec:derouting}

Recent MoE style LoRA variants enhance parameter efficiency by introducing multiple parallel low-rank branches together with a non-linear routing mechanism.
Given an input activation $\mathbf{x} \in \mathbb{R}^{n}$ and a stochastic perturbation $\mathbf{z}$ (e.g., induced by dropout), the adapted layer output is commonly formulated as
\begin{equation}
    f(\mathbf{x}, \mathbf{z}) = \mathbf{W} \mathbf{x}
    + \sum_{i=1}^{N} \omega_i(\mathbf{x}, \mathbf{z}) \, \mathbf{B}_i \mathbf{A}(\mathbf{x} \odot \mathbf{z}),
    \label{eq:moe_lora}
\end{equation}

As established in Section~\ref{sec:theoretical}, the non-linearity of $\omega_i$ induces an intrinsic
\emph{training--inference discrepancy}: during training, stochastic routing and dropout distribute gradients across distinct computational paths, whereas inference aggregates these paths through a deterministic non-linear mixture.
This mismatch results in biased estimates of the expected adaptation and compromises representation stability in multi-task settings.

To address this issue, we adopt a \textbf{de-routing} strategy that removes the routing mechanism entirely.
All LoRA heads are treated as contributing uniformly, yielding
\begin{equation}
    \hat{f}(\mathbf{x}, \mathbf{z}) = \mathbf{W} \mathbf{x}
    + \sum_{i=1}^{N} \mathbf{B}_i \mathbf{A}(\mathbf{x} \odot \mathbf{z}).
    \label{eq:derouted_lora}
\end{equation}
This transformation collapses the MoE-style architecture into a parallel multi-head LoRA without gating, ensuring that training and inference share an identical forward computation graph.

Importantly, de-routing eliminates the need for explicit task partitioning.
Instead, all heads operate jointly on a shared low-rank representation, yielding a unified parameterization that prioritizes common structure over fragmented, task-specific subspaces (see Section~\ref{sec:B_Similarity} for illustration).

\subsection{Consistency-Driven Alignment in LoRA-A}
\label{sec:consistency}

\paragraph{Hessian-based Instability Analysis.}
While the de-routing paradigm simplifies the architecture, Equation~\eqref{eq:gap_bound} suggests that the training--inference discrepancy (TID) remains inherently unoptimized during standard training. To characterize this instability, let $g(\mathbf{z}) = \mathbf{A}(\mathbf{x} \odot \mathbf{z})$. We perform a second-order Taylor expansion of $g(\mathbf{z})$ around the expectation $\bar{\mathbf{z}}$:
\begin{equation}
    \mathbb{E}_{\mathbf{z}}[g(\mathbf{z})] - g(\bar{\mathbf{z}}) \approx \frac{1}{2}
    \mathrm{Tr}\!\left(
    \nabla^{2} g(\bar{\mathbf{z}}) \cdot \mathrm{Var}(\mathbf{z})
    \right).
    \label{eq:hessian_gap}
\end{equation}
where $\nabla^{2} g(\bar{\mathbf{z}})$ is the Hessian matrix representing the curvature of the LoRA-A projection space. The detailed derivation of Equation~\eqref{eq:hessian_gap} is provided in Appendix~\ref{app:taylor}.

Equation~\eqref{eq:hessian_gap} reveals that the TID is primarily driven by the curvature of the latent manifold; high sensitivity to dropout perturbations $\mathbf{z}$ manifests as a large Hessian norm, leading to inference instability. 

\paragraph{From Curvature Smoothing to Consistency.}
Directly optimizing the Hessian is computationally prohibitive. However, inspired by \textit{consistency training} \citep{R-drop}, we observe that regularizing the variance of latent representations under stochastic perturbations serves as a first-order proxy for smoothing the Hessian. Specifically, if we model the latent representations $\mathbf{h}^{(1)}, \mathbf{h}^{(2)}$ from two independent forward passes as isotropic Gaussians $\mathcal{P} \sim \mathcal{N}(\boldsymbol{\mu}, \sigma^2 \mathbf{I})$, the symmetric Kullback-Leibler (KL) divergence simplifies to a squared $\ell_2$ distance:
\begin{equation}
    D_{\mathrm{KL}}(\mathcal{P}_1 \| \mathcal{P}_2) = \frac{1}{2\sigma^2} \|\mathbf{h}^{(1)} - \mathbf{h}^{(2)}\|_2^2,
\end{equation}
where $\mathbf{h}^{(j)} = \mathbf{A}(\mathbf{x} \odot \mathbf{z}^{(j)})$ represents the $j$-th stochastic projection. Minimizing this divergence effectively constrains the spectral norm of $\nabla^2 g(\bar{\mathbf{z}})$, fostering a flatter representation manifold.

\paragraph{The CD-LoRA Objective.}
Building on the curvature-smoothing intuition, we propose the Consistency-Driven Alignment (CD-A) mechanism. For each input $\mathbf{x}$, we perform two independent forward passes through the shared LoRA-A matrix using distinct dropout masks $\mathbf{z}^{(1)}$ and $\mathbf{z}^{(2)}$, yielding two latent realizations $\mathbf{h}^{(1)} = g(\mathbf{z}^{(1)})$ and $\mathbf{h}^{(2)} = g(\mathbf{z}^{(2)})$. To minimize the training--inference discrepancy, we define the symmetric KL divergence as $D_{\mathrm{SKL}}(P\|Q) = D_{\mathrm{KL}}(P\|Q) + D_{\mathrm{KL}}(Q\|P)$. The consistency loss is then simplified as:
\begin{equation}
    \mathcal{L}_{\mathrm{cons}} = \frac{1}{2} D_{\mathrm{SKL}}\big(P(\cdot | \mathbf{x}, \mathbf{h}^{(1)}) \,\|\, P(\cdot | \mathbf{x}, \mathbf{h}^{(2)})\big),
    \label{eq:consistency_loss}
\end{equation}
\textit{\textbf{Notations:} $\mathbf{h}^{(j)}$ denotes the $j$-th latent realization from LoRA-A; $P(\cdot | \mathbf{x}, \mathbf{h}^{(j)})$ is the corresponding output distribution.}

The final training objective of CD-LoRA integrates the task-specific negative log-likelihood (NLL) with the consistency regularization:
\begin{equation}
    \mathcal{L}_{\mathrm{total}} = \mathcal{L}_{\mathrm{task}} + \lambda \, \mathcal{L}_{\mathrm{cons}},
    \label{eq:total_loss}
\end{equation}
\textit{\textbf{Notations:} $\lambda$ is a balancing hyperparameter governing the intensity of the consistency constraint; $\mathcal{L}_{\mathrm{task}}$ denotes the primary empirical risk across the multi-task suite, ensuring task-specific predictive proficiency; $\mathcal{L}_{\mathrm{cons}}$ serves as a structural regularizer to minimize the Hessian-driven training--inference discrepancy.}

To facilitate understanding, we summarize the complete derivation of CD-LoRA in Appendix~\ref{app:derivation}. 

\begin{figure}[t]
    \centering
    \includegraphics[width=0.9\linewidth]{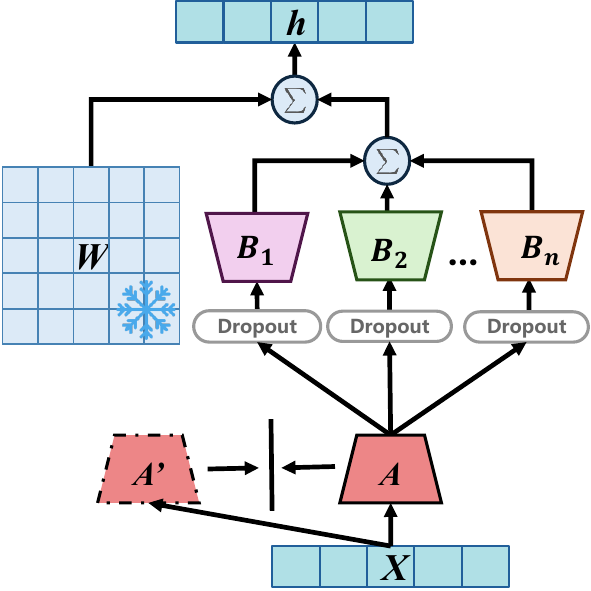}
    \caption{Overview of the CD-LoRA framework.}
    \label{fig:overview}
\end{figure}

Since the input $\mathbf{x}$ is shared across both passes, the consistency loss $\mathcal{L}_{\mathrm{cons}}$ explicitly aligns representations within the LoRA-$\mathbf{A}$ module, regularizing the curvature of the shared low-rank manifold across tasks. This alignment improves robustness to stochastic perturbations during training and effectively closes the training--inference gap. As illustrated in Figure~\ref{fig:overview}, CD-LoRA removes unstable routers and adopts a dual-pass design: $\mathbf{A}$ serves as the primary projection, while $\mathbf{A}'$ provides an auxiliary stochastic view. The consistency loss enforces agreement between these views, smoothing the representation manifold and stabilizing inference.

\section{Experiments}
\label{sec:experiments}

We evaluate CD-LoRA against representative LoRA-based multi-task
fine-tuning methods. Unless otherwise specified, all implementation details and hyperparameter
settings are deferred to the Appendix~\ref{app:Main_Experiment}.

\subsection{Experimental Setup}

\paragraph{Backbone Models.}
We conduct experiments on two decoder-only LLMs,
\textbf{Qwen2.5-7B} and \textbf{Qwen2.5-14B}~\citep{qwen2.5},
representing mid-scale and large-scale backbones, respectively.

\paragraph{Baselines.}
We compare CD-LoRA with widely used parameter-efficient multi-task
adaptation methods, including \textbf{LoRA}~\citep{lora}, \textbf{HydraLoRA}~\citep{Hydralora},
\textbf{R-LoRA}~\citep{R-LoRA}, and \textbf{M-LoRA}~\cite{align}.
All methods are implemented under identical optimization settings,
batch sizes, and training schedules to ensure a fair comparison.

\paragraph{Evaluation Protocol.}
All models are jointly fine-tuned and evaluated on a unified benchmark comprising the 12 diverse natural language understanding (NLU) tasks listed in Appendix~\ref{app:Twelve-Task}, covering natural language inference, commonsense reasoning, sentiment analysis, and mathematical reasoning. Task performance is reported using standard accuracy, and overall performance is summarized by the average across tasks.

\subsection{Main Results on Multi-Task Benchmarks}

Table~\ref{tab:main_results_final} reports the performance of all methods.
\begin{figure}[t]
    \centering
    \includegraphics[width=1.0\linewidth]{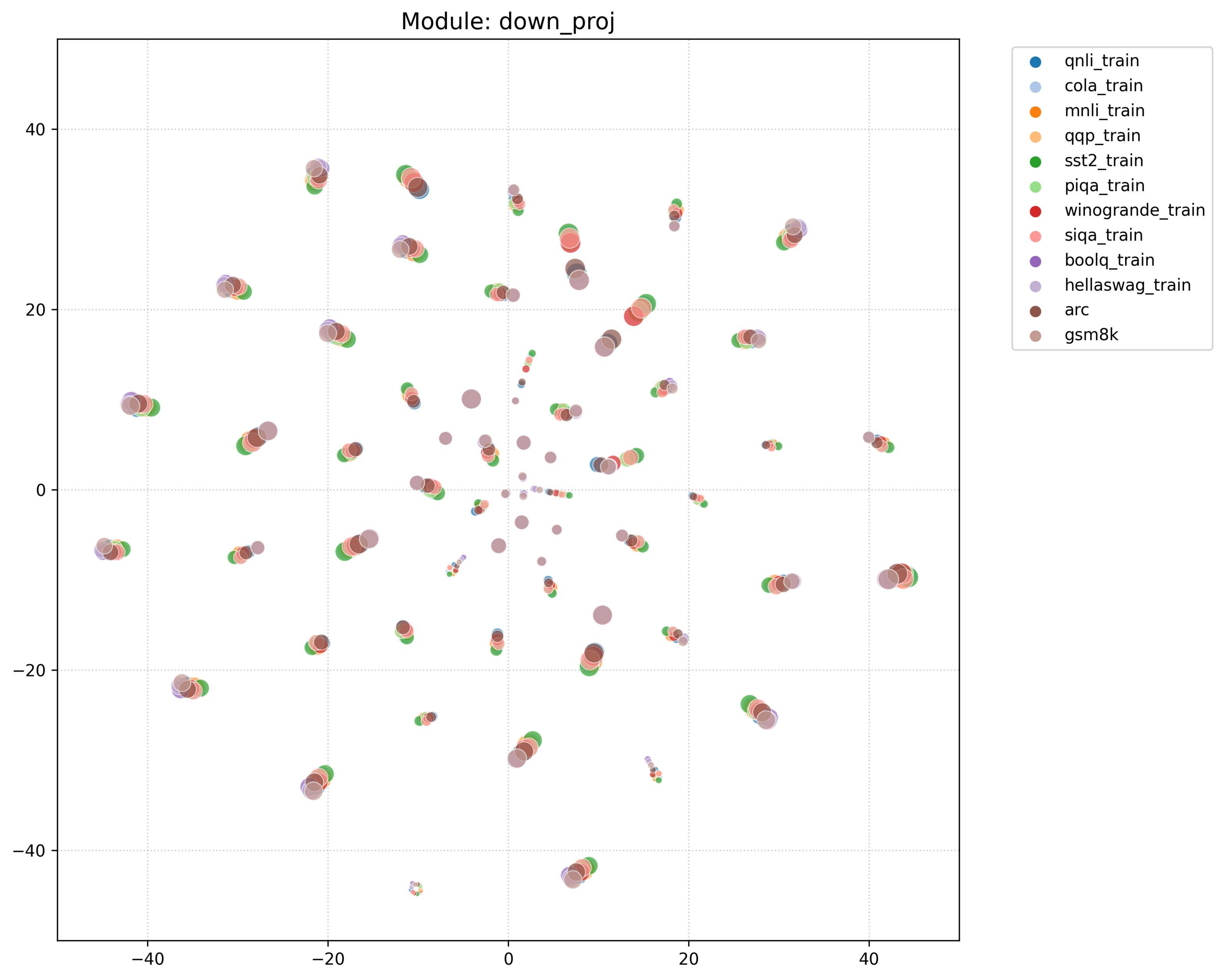}
    \caption{
t-SNE visualization of LoRA-A activations learned by CD-LoRA.
Different colors correspond to different tasks.
CD-LoRA exhibits compact and well-separated clusters, indicating improved task-level representation consistency.
}
    \label{fig:tsne_cdlora}
\end{figure}
\begin{table*}[t]
\centering
\caption{Performance comparison of multi-task fine-tuning methods on
Qwen2.5-7B and Qwen2.5-14B.}
\setlength{\tabcolsep}{6pt}
\resizebox{0.95\textwidth}{!}{
\begin{tabular}{lcccccccccccccc}
\toprule
\textbf{Method} &
\textbf{QNLI} & \textbf{CoLA} & \textbf{MNLI} & \textbf{QQP} &
\textbf{SST-2} & \textbf{PIQA} & \textbf{Wino} & \textbf{SIQA} &
\textbf{BoolQ} & \textbf{Hella} & \textbf{ARC} & \textbf{GSM8K} &
\textbf{Avg.} & \textbf{Mem{$\ast$}} \\
\midrule
\multicolumn{15}{l}{\textit{Qwen2.5-7B}} \\
\midrule
LoRA      & 82.40 & 84.30 & 87.20 & 86.40 & 95.44 & 88.30 & 78.70 & 60.30 & 88.10 & 94.20 & 91.34 & 48.50 & 82.10 & 60062 \\
HydraLoRA & 83.70 & 85.20 & 88.10 & 87.30 & 96.33 & 89.70 & 79.20 & 60.70 & 88.60 & 95.30 & 91.60 & \textbf{51.60} & 83.11 & 87860 \\
R-LoRA    & 83.70 & 85.40 & 88.50 & 86.90 & 96.44 & 90.10 & 79.20 & 61.00 & 88.90 & 95.20 & 91.83 & 50.30 & 83.12 & 66070 \\
M-LoRA    & 84.00 & 85.60 & 88.60 & 87.20 & 96.56 & 90.20 & 80.90 & 60.40 & 89.20 & \textbf{95.50} & \textbf{92.29} & 50.70 & 83.43 & 49868 \\
\textbf{CD-LoRA} & \textbf{84.60} & \textbf{85.70} & \textbf{89.60} & \textbf{87.80} &
\textbf{96.79} & \textbf{90.20} & \textbf{83.50} & \textbf{61.40} &
\textbf{89.60} & 95.30 & 92.17 & 50.40 & \textbf{83.92} & 56882 \\
\midrule
\multicolumn{15}{l}{\textit{Qwen2.5-14B}} \\
\midrule
LoRA      & 83.50 & 85.90 & 88.70 & 87.30 & 96.33 & 90.40 & 79.50 & 60.90 & 89.00 & 95.70 & 92.56 & 51.60 & 83.45 & 73691 \\
HydraLoRA & 85.10 & 86.40 & 89.40 & 86.80 & 96.63 & 91.70 & 85.70 & 61.10 & 90.37 & 96.40 & 93.21 & 53.30 & 84.68 & 107796 \\
R-LoRA    & 85.20 & 86.40 & \textbf{89.80} & 86.40 & 96.67 & 91.77 & 84.70 & 61.00 & 90.71 & 96.80 & 93.33 & \textbf{53.90} & 84.72 & 81062 \\
M-LoRA    & 86.30 & 86.60 & 89.30 & 86.70 & \textbf{97.13} & 92.31 & 85.90 & 61.80 & 91.26 & 96.80 & 92.98 & 53.60 & 85.06 & 65736 \\
\textbf{CD-LoRA} & \textbf{87.30} & \textbf{87.30} & 89.50 & \textbf{87.40} &
96.67 & \textbf{92.63} & \textbf{89.60} & \textbf{63.90} &
\textbf{91.59} & \textbf{97.10} & \textbf{93.90} & \textbf{53.90} &
\textbf{85.90} & 74870 \\
\bottomrule
\end{tabular}
}
\label{tab:main_results_final}
\end{table*}

\stepcounter{footnote}
\footnotetext{$^{\ast}$ \textbf{Mem}: Peak GPU memory usage (MB) measured under identical hardware and training configurations.}

\paragraph{Discussion.}
Experimental results across both backbones demonstrate that CD-LoRA consistently achieves superior average performance while maintaining competitive memory efficiency. 

On the Qwen2.5-7B backbone, CD-LoRA achieves the highest \textbf{Avg.} score of \textbf{83.92\%}, outperforming the routing-based R-LoRA (83.12\%) by \textbf{0.80\%} and the strongest router-free baseline M-LoRA (83.43\%) by \textbf{0.49\%}. Notably, while achieving superior performance, CD-LoRA's peak memory usage (56,882 MB) is significantly lower than that of routing-based HydraLoRA (87,860 MB), representing a \textbf{35.3\% reduction} in memory overhead. The trend scales positively with model capacity: on Qwen2.5-14B, CD-LoRA further widens the gap, reaching \textbf{85.90\% Avg.}, a \textbf{1.18\% improvement} over R-LoRA. 

Furthermore, the consistent gains across diverse tasks—particularly the significant improvements in complex reasoning tasks like Wino~\citep{Winogrande} and SIQA~\citep{SIQA}—indicate that consistency-driven alignment effectively complements the de-routed architecture by fostering a more stable and task-agnostic representation manifold.

\subsection{Analysis of LoRA-A Subspace Consistency}
\label{sec:LoRA-A}
Under the same conditions as in Section~\ref{sec:Analysis}, we visualize the LoRA-A representations of CD-LoRA, as shown in Figures~\ref{fig:tsne_cdlora} and~\ref{fig:tsne_comparison}. CD-LoRA yields significantly more compact and task-separable embeddings, indicating that enforcing consistency in the shared low-rank space promotes stable and discriminative task representations—validating the effectiveness of our consistency-driven approach.


\section{Ablation Studies}
\begin{table*}[t]
\centering
\caption{
Module-level ablation on five NLU tasks using 7B models.
}
\setlength{\tabcolsep}{8pt}
\resizebox{0.75\textwidth}{!}{
\begin{tabular}{lcc ccccc cc}
\toprule
\textbf{Method} 
& \textbf{Router} 
& \textbf{Consistency} 
& \textbf{QNLI} 
& \textbf{MNLI} 
& \textbf{Wino} 
& \textbf{BoolQ} 
& \textbf{ARC} 
& \textbf{Avg.} 
& \textbf{Mem{$\ast$}} \\
\midrule
R-LoRA\textsuperscript{$\dagger$}   
& \cmark & \xmark 
& 84.30 & 90.30 & 83.00 & 88.40 & 91.94 & 87.59 & 66071 \\

CR-LoRA\textsuperscript{$\ddagger$}  
& \cmark & \cmark 
& 84.20 & 90.20 & 83.90 & 89.10 & 92.17 & 87.91 & 80943 \\

DR-LoRA\textsuperscript{$\ast$}  
& \xmark  & \xmark  
& 85.10 & 90.60 & 84.60 & 89.00 & 92.17 & 88.29 & 49869 \\

\textbf{CD-LoRA}\textsuperscript{$\S$} 
& \xmark & \cmark 
& \textbf{85.40} & \textbf{90.90} & \textbf{85.00} & \textbf{89.20} & \textbf{92.29} & \textbf{88.56} & 56883 \\
\bottomrule
\end{tabular}
}
\vspace{1mm}
\label{tab:ablation}
\end{table*}
\stepcounter{footnote}
\footnotetext{$^{\dagger}$ \textbf{R-LoRA}: Baseline multi-task LoRA using non-linear routing mechanisms.}
\stepcounter{footnote}
\footnotetext{$^{\ddagger}$ \textbf{CR-LoRA}: R-LoRA variant incorporating our consistency alignment module without removing the router.}
\stepcounter{footnote}
\footnotetext{$^{\ast}$ \textbf{DR-LoRA}: \textbf{D}e-\textbf{R}outed LoRA, a simplified parallel multi-head structure without routing or consistency constraints.}
\stepcounter{footnote}
\footnotetext{$^{\S}$ \textbf{CD-LoRA}: Our proposed \textbf{C}onsistency-\textbf{D}riven LoRA, combining the de-routed architecture with explicit consistency alignment.}

\subsection{Module-level Ablation on 7B Models}
To disentangle the individual and combined effects of routing mechanisms and consistency alignment, we conduct a controlled module-level ablation on the Qwen2.5-7B backbone across five representative NLU tasks. As summarized in Table~\ref{tab:ablation}, we construct a $2\times2$ ablation matrix by independently enabling or disabling (i) dynamic routing and (ii) explicit consistency alignment.

\paragraph{Efficacy of De-routing.} 
Our results reveal that simply removing the dynamic routing mechanism (DR-LoRA vs.\ R-LoRA) yields a significant performance boost, elevating the \textbf{Avg.} score from \textbf{87.59\%} to \textbf{88.29\%} while simultaneously reducing peak memory usage by \textbf{24.5\%} (from 66,071 MB to 49,869 MB). This substantial gain suggests that routing-induced non-linearity introduces detrimental optimization overhead rather than task-specific benefits. This empirical evidence directly supports our theoretical formalization in Section~\ref{sec:theoretical}, confirming that the non-linear training--inference discrepancy serves as a performance bottleneck.

\paragraph{Synergy between Alignment and De-routing.} 
The introduction of consistency-driven alignment further enhances the de-routed architecture. \textbf{CD-LoRA} (the full model) achieves the highest performance across all evaluated tasks, reaching an \textbf{Avg.} of \textbf{88.56\%}. Notably, while adding consistency alignment to a routing-based model (CR-LoRA) provides a marginal improvement of \textbf{0.32\%} over R-LoRA, it incurs a significant \textbf{22.5\%} memory penalty. In contrast, CD-LoRA achieves superior results with far greater efficiency, indicating that consistency regularization is most effective when the structural instability of routers is eliminated.

To further validate the effectiveness and generalizability of our method, we conduct additional experiments; details of the setup and results are provided in Appendix~\ref{app:BBH_setup}.

\subsection{Mechanistic Analysis: Subspace Coherence in LoRA-B}
\label{sec:B_Similarity}

To uncover the mechanistic origins of CD-LoRA's robust multi-task generalization, we conduct a fine-grained analysis of the learned LoRA-B matrices on the Qwen2.5-14B backbone. At this scale, representational divergence across tasks and functional modules becomes particularly pronounced. \\
We focus on three critical Transformer components:\\ \texttt{up\_proj}:   Expands input features into a high-dimensional latent space.\\  
\texttt{gate\_proj}: Modulates feature flow via nonlinear gating.\\  
\texttt{down\_proj}: Projects the gated representations back to the residual dimension.

We quantify the coherence of the learned low-rank subspaces by calculating the average pairwise cosine similarity $\mathcal{S}$ between task-specific LoRA-$\mathbf{B}$ . For any two tasks $i$ and $j$ in a given layer, the similarity is defined as:
\begin{equation}
    \text{cos}(\mathbf{B}_i, \mathbf{B}_j) = \frac{\text{vec}(\mathbf{B}_i)^{\top} \text{vec}(\mathbf{B}_j)}{\|\text{vec}(\mathbf{B}_i)\|_2 \|\text{vec}(\mathbf{B}_j)\|_2},
    \label{eq:cosine_sim}
\end{equation}
where $\text{vec}(\cdot)$ denotes the flattening operation and $\|\cdot\|_2$ is the $\ell_2$ norm. To evaluate the global alignment of a functional module (e.g., \texttt{gate\_proj}), we report the mean similarity $\bar{\mathcal{S}}$ across all $L$ layers:
\begin{equation}
    \bar{\mathcal{S}} = \frac{1}{L} \sum_{l=1}^{L} \left( \frac{2}{K(K-1)} \sum_{1 \le i < j \le K} \text{cos}(\mathbf{B}_{l,i}, \mathbf{B}_{l,j}) \right),
    \label{eq:avg_sim}
\end{equation}
where $L=48$ for the 14B model and $K$ denotes the number of heads in the multi-head architecture.

\paragraph{Quantitative Analysis.}
As summarized in Table~\ref{tab:bi_similarity_refined}, routing-based methods (R-LoRA and CR-LoRA) exhibit significant \textit{subspace fragmentation}. R-LoRA achieves a mean similarity of only \textbf{77.64\%}, dropping further to \textbf{74.80\%} in the \texttt{gate\_proj}. This suggests that non-linear routing prompts tasks to converge into isolated, heterogeneous parameter regions. While de-routing (DR-LoRA) increases the average similarity to \textbf{82.07\%} by enforcing shared parameterization, it still lacks explicit cross-task coordination.

Notably, CD-LoRA achieves the highest similarity consistently across all projections, with a global average of \textbf{84.93\%}. In the most challenging \texttt{gate\_proj}, CD-LoRA improves the similarity from R-LoRA's 74.80\% to \textbf{84.49\%} (\textbf{+9.69\%} absolute gain). This uniform enhancement indicates that consistency-driven alignment effectively stabilizes the representation manifold across diverse Transformer operators.

\begin{figure}[t]
\centering
\includegraphics[width=1.0\linewidth]{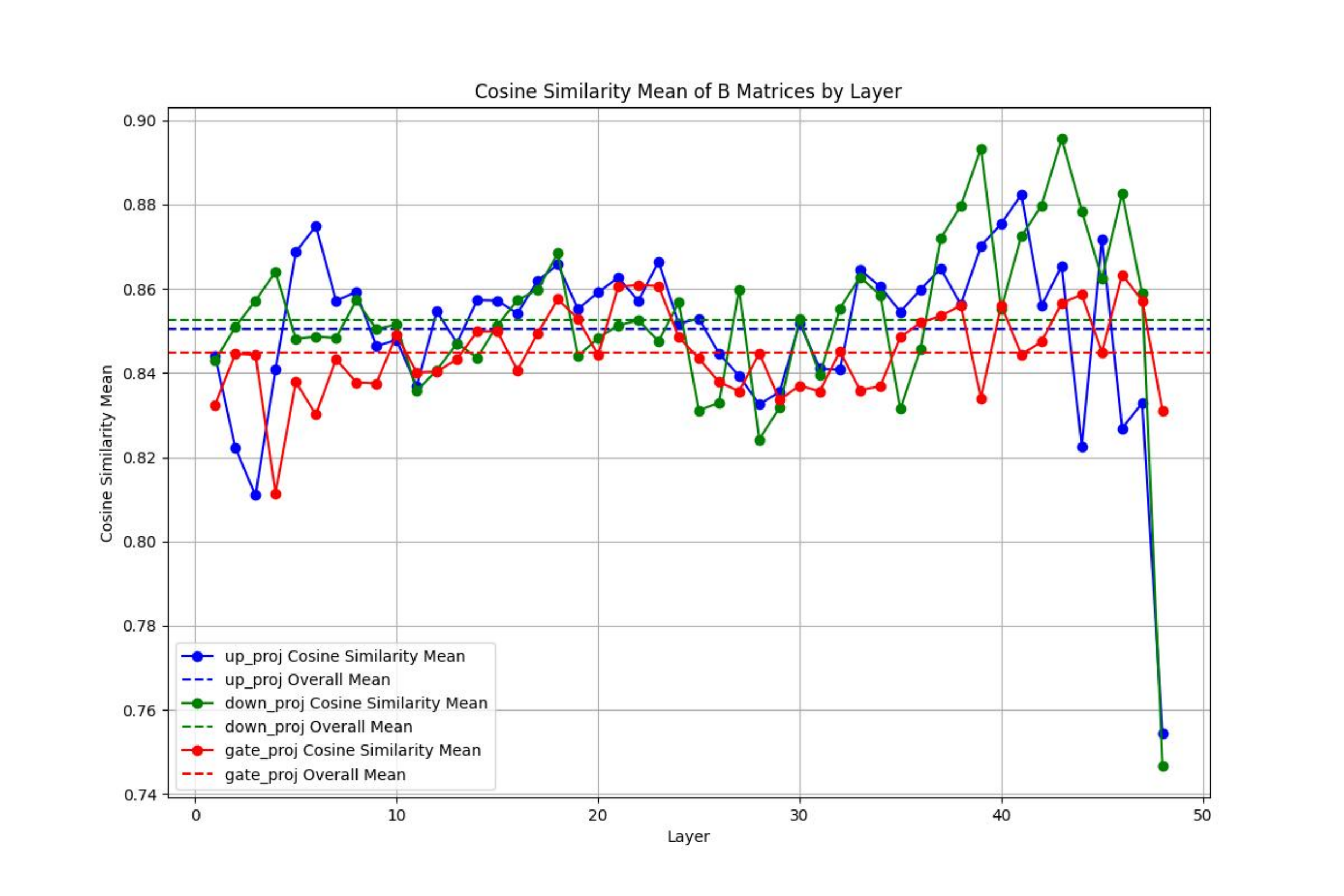}
\caption{Layer-wise $\mathbf{B}$ matrix similarity heatmap for CD-LoRA. The compact and uniform patterns across all projections validate the efficacy of consistency-driven alignment.}
\label{fig:CD-LoRA_sim}
\end{figure}

\paragraph{Qualitative Visualization.}
Figure~\ref{fig:CD-LoRA_sim} illustrates the layer-wise cosine similarity of CD-LoRA across different projection layers. In contrast to the erratic fluctuations or fragmented representations typical of routing-based approaches (R-LoRA in Figure~\ref{fig:R-LoRA_sim} and CR-LoRA in Figure~\ref{fig:CR-LoRA_sim}), CD-LoRA maintains a high and stable similarity mean across nearly all layers. This cross-layer consistency qualitatively reinforces our findings: the consistency-driven alignment mechanism effectively constrains the low-rank subspaces of diverse tasks within a coherent manifold, thereby preventing representational drift during multi-task adaptation in large-scale models.

\paragraph{Discussion.}
These results provide direct mechanistic evidence that the performance gains of CD-LoRA stem from \textbf{subspace coherence}. By eliminating routing-induced fragmentation and explicitly aligning task representations, CD-LoRA enforces a shared inductive bias that prevents weight drifting. This structural stability becomes increasingly critical as model capacity and task complexity scale.

\begin{table}[htbp]
\centering
\caption{Quantification of $\mathbf{B}$ matrix similarity (\%) across projection layers for Qwen2.5-14B.}
\setlength{\tabcolsep}{10pt}
\resizebox{\columnwidth}{!}{
\begin{tabular}{lcccc}
\toprule
\textbf{Method} & \textbf{up\_proj} & \textbf{down\_proj} & \textbf{gate\_proj} & \textbf{Avg.} \\
\midrule
R-LoRA\textsuperscript{$\dagger$}  & 77.98 & 80.15 & 74.80 & 77.64 \\
CR-LoRA\textsuperscript{$\ddagger$} & 79.30 & 82.00 & 75.23 & 78.84 \\
DR-LoRA\textsuperscript{$\ast$} & 82.20 & 82.50 & 81.52 & 82.07 \\
\textbf{CD-LoRA}\textsuperscript{$\S$} & \textbf{85.03} & \textbf{85.26} & \textbf{84.49} & \textbf{84.93} \\
\bottomrule
\end{tabular}
}
\label{tab:bi_similarity_refined}
\end{table}

\subsection{Sensitivity to the Regularization Weight $\lambda$}
\label{sec:lambda_sensitivity}

The hyperparameter $\lambda$ in Equation~\eqref{eq:total_loss} governs the trade-off between the task-specific empirical risk $\mathcal{L}_{\mathrm{task}}$ and the consistency regularizer $\mathcal{L}_{\mathrm{cons}}$. We evaluate this sensitivity using the Qwen2.5-3B backbone across the 12-task in~\ref{app:Twelve-Task}. 

\begin{figure}[t]
    \centering
    \includegraphics[width=0.9\linewidth]{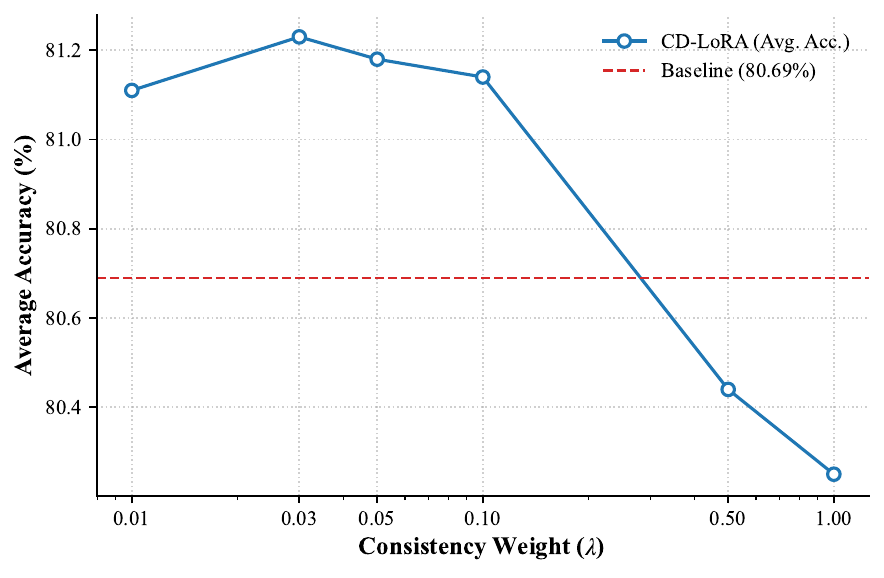}
    \caption{Sensitivity analysis of the regularization weight $\lambda$ on Qwen2.5-3B. The model maintains superior performance over the baseline within a wide range of $\lambda$, with optimal multi-task accuracy achieved at $\lambda = 0.03$.}
    \label{fig:lambda_sensitivity}
\end{figure}

As illustrated in Figure~\ref{fig:lambda_sensitivity}, CD-LoRA demonstrates strong robustness within the range $\lambda \in [0.01, 0.1]$, consistently yielding an average accuracy above \textbf{81.1\%}. Performance peaks at $\lambda = 0.03$ (\textbf{81.23\%}), providing a \textbf{0.54\%} absolute gain over the baseline (\textbf{80.69\%}). Notably, when $\lambda$ is further increased to $0.5$ or $1.0$, the model’s performance exhibits a mild downward trend, with accuracy values of $80.44\%$ and $80.25\%$, respectively, which are slightly below the baseline level.

This suggests that while moderate consistency alignment stabilizes the shared LoRA-$\mathbf{A}$ manifold, excessive regularization (large $\lambda$) over-constrains the low-rank projection space, impairing the model’s ability to retain task-specific nuances and degrading multi-task performance.

\section{Conclusion}
This work provides a rigorous theoretical examination of router-based multi-head LoRA, identifying a critical training-inference discrepancy caused by nonlinear gating that undermines stability and task decoupling. To resolve this, we introduce \textbf{Consistency-Driven Low-Rank Adaptation (CD-LoRA)}, which replaces complex routing with explicit consistency alignment in a shared low-rank space. This design eliminates the structural gap between phases and enhances task separability, as demonstrated by superior t-SNE clustering patterns. 

Empirical results show that CD-LoRA consistently outperforms superior baselines like R-LoRA with lower computational overhead. Our findings suggest that collaborative learning within a robust shared space offers better generalization than traditional structural isolation. By challenging the necessity of architectural modularity in multi-task scenarios, this work establishes a streamlined and effective paradigm for robust parameter-efficient fine-tuning (PEFT).

\section*{Impact Statement}

This paper presents work whose goal is to advance the field of Machine
Learning. There are many potential societal consequences of our work, none
which we feel must be specifically highlighted here.

\nocite{langley00}

\bibliography{example_paper}
\bibliographystyle{icml2026}

\newpage
\appendix
\onecolumn
\appendix
\onecolumn
\section{Derivation of CD-LoRA}
\label{app:derivation}

In this appendix, we provide a complete derivation of the training--inference discrepancy in routing-based LoRA architectures and its mitigation via consistency-driven alignment. All steps are explicitly stated for clarity and reproducibility.

\subsection{Preliminaries and Notation}

Let $\mathbf{x} \in \mathbb{R}^n$ denote the input activation to a linear layer, and let
$\mathbf{z} \sim p(\mathbf{z})$ be a stochastic perturbation induced by dropout, with
expectation $\bar{\mathbf{z}} = \mathbb{E}[\mathbf{z}]$ and covariance
$\mathrm{Var}(\mathbf{z}) = \mathbb{E}[(\mathbf{z}-\bar{\mathbf{z}})(\mathbf{z}-\bar{\mathbf{z}})^{\top}]$.

In routing-based LoRA variants, the low-rank update is defined as
\begin{equation}
    \Delta \mathbf{W}(\mathbf{x}, \mathbf{z})
    = \sum_{i=1}^{N} \omega_i(\mathbf{x}, \mathbf{z}) \,
    \mathbf{B}_i \mathbf{A}(\mathbf{x} \odot \mathbf{z}),
    \label{eq:app_routing}
\end{equation}
where $\mathbf{A} \in \mathbb{R}^{r \times n}$ is the shared down-projection,
$\mathbf{B}_i \in \mathbb{R}^{m \times r}$ are head-specific up-projections, and
$\omega_i(\cdot)$ are non-linear routing weights produced by a softmax-based gating network.

At inference time, the stochastic mask is replaced by its expectation $\bar{\mathbf{z}}$.
Due to the non-linearity of $\omega_i(\cdot)$, the expected training-time update generally
differs from the inference-time update, yielding the training--inference discrepancy.

\subsection{Router-Free Decomposition}

To isolate the effect of stochastic projection, we consider a router-free multi-head LoRA
variant with uniform weights. The discrepancy for a fixed input $\mathbf{x}$ is
\begin{equation}
    \mathrm{TID}(\mathbf{x}) =
    \left\|
    \frac{1}{N} \sum_{i=1}^{N} \mathbf{B}_{i}
    \left(
    \mathbb{E}_{\mathbf{z}}[\mathbf{A}(\mathbf{x} \odot \mathbf{z})]
    - \mathbf{A}(\mathbf{x} \odot \bar{\mathbf{z}})
    \right)
    \right\|.
    \label{eq:app_tid}
\end{equation}
By sub-multiplicativity of matrix norms, this yields the upper bound
\begin{equation}
    \mathrm{TID}(\mathbf{x}) \le
    \left( \frac{1}{N} \sum_{i=1}^{N} \|\mathbf{B}_{i}\| \right)
    \cdot
    \left\|
    \mathbb{E}_{\mathbf{z}}[\mathbf{A}(\mathbf{x} \odot \mathbf{z})]
    - \mathbf{A}(\mathbf{x} \odot \bar{\mathbf{z}})
    \right\|.
    \label{eq:app_tid_bound}
\end{equation}
Thus, the discrepancy is entirely governed by the stability of the shared projection
$\mathbf{A}$ under stochastic perturbations.

\subsection{Second-Order Taylor Expansion}
\label{app:taylor}
Define the projection function
\begin{equation}
    g(\mathbf{z}) = \mathbf{A}(\mathbf{x} \odot \mathbf{z}).
\end{equation}
We perform a second-order Taylor expansion of $g(\mathbf{z})$ around
$\bar{\mathbf{z}} = \mathbb{E}[\mathbf{z}]$:
\begin{equation}
\begin{aligned}
    g(\mathbf{z})
    \approx\;
    & g(\bar{\mathbf{z}})
    + \nabla g(\bar{\mathbf{z}})^{\top} (\mathbf{z} - \bar{\mathbf{z}}) \\
    & + \frac{1}{2}
    (\mathbf{z} - \bar{\mathbf{z}})^{\top}
    \nabla^{2} g(\bar{\mathbf{z}})
    (\mathbf{z} - \bar{\mathbf{z}}).
\end{aligned}
\label{eq:app_taylor}
\end{equation}

Taking expectations with respect to $\mathbf{z}$ term by term:

\paragraph{First term.}
\begin{equation}
    \mathbb{E}_{\mathbf{z}}[g(\bar{\mathbf{z}})] = g(\bar{\mathbf{z}}).
\end{equation}

\paragraph{Second term.}
\begin{equation}
\begin{aligned}
    \mathbb{E}_{\mathbf{z}}\!\left[
    \nabla g(\bar{\mathbf{z}})^{\top} (\mathbf{z} - \bar{\mathbf{z}})
    \right]
    &= \nabla g(\bar{\mathbf{z}})^{\top}
    \mathbb{E}_{\mathbf{z}}[\mathbf{z} - \bar{\mathbf{z}}] \\
    &= \nabla g(\bar{\mathbf{z}})^{\top}
    (\mathbb{E}[\mathbf{z}] - \bar{\mathbf{z}}) \\
    &= 0.
\end{aligned}
\end{equation}

\paragraph{Third term.}
\begin{equation}
\begin{aligned}
    \mathbb{E}_{\mathbf{z}}\!\left[
    \frac{1}{2}
    (\mathbf{z} - \bar{\mathbf{z}})^{\top}
    \nabla^{2} g(\bar{\mathbf{z}})
    (\mathbf{z} - \bar{\mathbf{z}})
    \right]
    &=
    \frac{1}{2}
    \mathbb{E}_{\mathbf{z}}\!\left[
    \mathrm{Tr}\!\left(
    \nabla^{2} g(\bar{\mathbf{z}})
    (\mathbf{z} - \bar{\mathbf{z}})
    (\mathbf{z} - \bar{\mathbf{z}})^{\top}
    \right)
    \right].
\end{aligned}
\end{equation}

Combining all terms yields
\begin{equation}
    \mathbb{E}_{\mathbf{z}}[g(\mathbf{z})] - g(\bar{\mathbf{z}})
    \;\approx\;
    \frac{1}{2}
    \mathrm{Tr}\!\left(
    \nabla^{2} g(\bar{\mathbf{z}}) \cdot \mathrm{Var}(\mathbf{z})
    \right).
    \label{eq:app_final_gap}
\end{equation}

\subsection{From Consistency Regularization to Curvature Control}

Directly minimizing the Hessian norm $\|\nabla^2 g(\bar{\mathbf{z}})\|$ is computationally prohibitive.
Instead, CD-LoRA adopts \emph{consistency regularization} as a tractable surrogate that implicitly constrains the curvature of the projection function $g(\mathbf{z})$.

Consider two independent stochastic perturbations $\mathbf{z}^{(1)}, \mathbf{z}^{(2)} \sim p(\mathbf{z})$,
and the corresponding latent representations
\begin{equation}
    \mathbf{h}^{(1)} = g(\mathbf{z}^{(1)}) = \mathbf{A}(\mathbf{x} \odot \mathbf{z}^{(1)}), \quad
    \mathbf{h}^{(2)} = g(\mathbf{z}^{(2)}) = \mathbf{A}(\mathbf{x} \odot \mathbf{z}^{(2)}).
\end{equation}

We model the induced latent distributions as isotropic Gaussians,
\begin{equation}
    \mathcal{P}_k \;=\; \mathcal{N}(\mathbf{h}^{(k)}, \sigma^2 \mathbf{I}), \quad k \in \{1,2\}.
\end{equation}
Under this assumption, the Kullback--Leibler divergence admits a closed-form expression:
\begin{equation}
    D_{\mathrm{KL}}(\mathcal{P}_1 \| \mathcal{P}_2)
    = \frac{1}{2\sigma^2} \| \mathbf{h}^{(1)} - \mathbf{h}^{(2)} \|_2^2.
\end{equation}
The symmetric KL divergence therefore reduces to
\begin{equation}
    D_{\mathrm{SKL}}(\mathcal{P}_1 \| \mathcal{P}_2)
    = \frac{1}{\sigma^2} \| \mathbf{h}^{(1)} - \mathbf{h}^{(2)} \|_2^2.
    \label{eq:app_skl_l2}
\end{equation}

\paragraph{Connection to Second-Order Sensitivity.}
To connect Eq.~\eqref{eq:app_skl_l2} with the curvature of $g(\mathbf{z})$, we expand the difference
$\mathbf{h}^{(1)} - \mathbf{h}^{(2)}$ around $\bar{\mathbf{z}}$ using the Taylor expansion in
Eq.~\eqref{eq:app_taylor}:
\begin{equation}
\begin{aligned}
    \mathbf{h}^{(1)} - \mathbf{h}^{(2)}
    &\approx
    \nabla g(\bar{\mathbf{z}})^{\top} (\mathbf{z}^{(1)} - \mathbf{z}^{(2)}) \\
    &\quad + \frac{1}{2}
    \Big[
    (\mathbf{z}^{(1)} - \bar{\mathbf{z}})^{\top}
    \nabla^{2} g(\bar{\mathbf{z}})
    (\mathbf{z}^{(1)} - \bar{\mathbf{z}})
    -
    (\mathbf{z}^{(2)} - \bar{\mathbf{z}})^{\top}
    \nabla^{2} g(\bar{\mathbf{z}})
    (\mathbf{z}^{(2)} - \bar{\mathbf{z}})
    \Big].
\end{aligned}
\end{equation}

Taking expectation over $\mathbf{z}^{(1)}, \mathbf{z}^{(2)}$, the first-order terms vanish due to
$\mathbb{E}[\mathbf{z}^{(k)} - \bar{\mathbf{z}}] = 0$.
The dominant contribution is governed by the second-order term, yielding
\begin{equation}
    \mathbb{E}\!\left[ \| \mathbf{h}^{(1)} - \mathbf{h}^{(2)} \|_2^2 \right]
    \;\propto\;
    \mathrm{Tr}\!\left(
    \nabla^{2} g(\bar{\mathbf{z}}) \, \mathrm{Var}(\mathbf{z})
    \right).
    \label{eq:app_cons_hessian}
\end{equation}

\paragraph{Interpretation.}
Equation~\eqref{eq:app_cons_hessian} establishes that minimizing the consistency loss
\begin{equation}
    \mathcal{L}_{\mathrm{cons}}
    \;\propto\;
    \mathbb{E}\!\left[ \| \mathbf{A}(\mathbf{x} \odot \mathbf{z}^{(1)}) -
    \mathbf{A}(\mathbf{x} \odot \mathbf{z}^{(2)}) \|_2^2 \right]
\end{equation}
implicitly suppresses the curvature of the projection function $g(\mathbf{z})$.
In other words, consistency regularization acts as a \emph{second-order smoothness prior}
on the shared LoRA-$\mathbf{A}$ projection, directly targeting the source of the
training--inference discrepancy identified in Eq.~\eqref{eq:app_final_gap}.

\subsection{Summary}

Putting everything together, the derivation shows that:
(i) the training--inference discrepancy originates from the Hessian of the shared projection;
(ii) routing amplifies this effect through additional non-linearities;
and (iii) CD-LoRA mitigates the discrepancy by enforcing stochastic consistency, which
serves as a tractable surrogate for curvature minimization.
This completes the theoretical justification of CD-LoRA.

\section{Additional Experimental Details}

This appendix provides additional details on datasets, training configurations,
and implementation specifics of CD-LoRA, complementing the main paper and
facilitating reproducibility.

\subsection{Experimental Datasets}
\subsubsection{Five-Task Setting}
\label{app:Five-Task}
\begin{itemize}
    \item \textbf{QNLI}~\cite{QNLI}: A question-answering formulation of natural language inference.
    \item \textbf{MNLI}~\cite{MNLI}: Multi-Genre Natural Language Inference covering diverse text domains.
    \item \textbf{Winogrande}~\cite{Winogrande}: An adversarial commonsense reasoning benchmark inspired by Winograd schemas.
    \item \textbf{BoolQ}~\cite{BoolQ}: Yes/No question answering over short passages.
    \item \textbf{ARC}~\cite{ARC}: The AI Reasoning Challenge, focusing on grade-school science questions.
\end{itemize}

These tasks jointly cover entailment, commonsense reasoning, and multi-choice
question answering, allowing us to examine cross-task interference under a
moderate training scale.

\subsubsection{Twelve-Task Setting}
\label{app:Twelve-Task}
\begin{itemize}
    \item \textbf{QNLI}~\cite{QNLI}: Question-answering Natural Language Inference.
    \item \textbf{CoLA}~\cite{CoLA}: Corpus of Linguistic Acceptability.
    \item \textbf{MNLI}~\cite{MNLI}: Multi-Genre Natural Language Inference.
    \item \textbf{QQP}~\cite{QNLI}: Quora Question Pairs.
    \item \textbf{SST-2}~\cite{SST2}: Stanford Sentiment Treebank (binary classification).
    \item \textbf{PIQA}~\cite{PIQA}: Physical Interaction Question Answering.
    \item \textbf{Winogrande}~\cite{Winogrande}: Commonsense coreference resolution.
    \item \textbf{SIQA}~\cite{SIQA}: Social Interaction Question Answering.
    \item \textbf{BoolQ}~\cite{BoolQ}: Boolean Questions.
    \item \textbf{HellaSwag}~\cite{HellaSwag}: Adversarial commonsense inference with long contexts.
    \item \textbf{ARC}~\cite{ARC}: AI Reasoning Challenge.
    \item \textbf{GSM8K}~\cite{GSM8K}: Grade School Math reasoning benchmark.
\end{itemize}

This setting significantly increases task heterogeneity, especially by
introducing mathematical reasoning (GSM8K) and fine-grained linguistic
acceptability judgments (CoLA), making it well suited for stress-testing
multi-adapter coordination.

\subsection{BBH Benchmark}
\label{app:BBH}

We additionally evaluate CD-LoRA on the BIG-Bench Hard (BBH) benchmark
\cite{BBH}, which consists of a curated subset of challenging tasks from
BIG-Bench that are known to be difficult for LLMs without
explicit reasoning supervision.

BBH focuses on tasks that require multi-step reasoning, compositional
generalization, and robust instruction following. The benchmark spans diverse
reasoning categories, including logical deduction, algorithmic reasoning,
commonsense inference, and symbolic manipulation, making it particularly
well suited for assessing the generalization and coordination ability of
parameter-efficient adaptation methods.

Following prior work, we adopt the standard BBH evaluation protocol and report
accuracy averaged across all selected tasks. For models trained in the
12-task multi-task setting, BBH evaluation is performed in a zero-shot manner
using task-specific prompts provided by the benchmark. No BBH data is used
during training.

In Table~\ref{tab:bbh_comparison_final}, we report both the average performance
across BBH tasks (BBH Avg.) and the overall accuracy aggregated over all
examples (BBH Ov.). Memory consumption is measured as the peak GPU memory usage
during training, reflecting the practical efficiency of different adaptation
methods.

\subsection{Visualization Setup for Representation Analysis}
\label{app:tsne_setup}

This section describes the experimental setup used for the representation
visualization in Section~\ref{sec:Analysis}.

\paragraph{Model and Training Setting.}
All representations are extracted from a Qwen2.5-14B backbone equipped with
multi-task LoRA adapters trained on a 12-task~\ref{app:Twelve-Task} mixture.
The LoRA configuration follows the main experimental setting, with rank
$r=4$ and adapters inserted into the feed-forward network (FFN) projection
layers.
Unless otherwise specified, all experiments are conducted in evaluation mode
without gradient updates.

\paragraph{Target Module.}
We focus on the LoRA-A matrix corresponding to the \texttt{down\_proj} module
in the FFN.
This module implements the shared low-rank down-projection and serves as the
common feature bottleneck across tasks.
As shown in Section~\ref{sec:theoretical}, instability in this shared projection
is the dominant source of the training--inference discrepancy, motivating its
use for representation analysis.

\paragraph{Activation Extraction.}
For each transformer layer, we register forward hooks on the LoRA-A
(\texttt{down\_proj}) module and extract its input activations.
Given an input sequence, token-level activations are averaged across the
sequence length to obtain a single representation vector per layer.
All representations are detached from the computation graph and collected
on the CPU for subsequent analysis.

\paragraph{Task Sampling.}
For each task in the 12-task~\ref{app:Twelve-Task} training set, we uniformly sample a small number
of input examples (one example per task in the main analysis) to avoid bias
toward high-resource tasks.
Representations are extracted from all transformer layers for each sampled
input.

\paragraph{Dimensionality Reduction and Visualization.}
The collected representations are projected to two dimensions using t-SNE
with PCA initialization and a fixed random seed for reproducibility.
All visualizations use identical axis limits and aspect ratios to ensure
consistent geometric interpretation across settings.
Different colors denote different tasks, while marker sizes encode layer depth.

\subsection{Main Experiment}
\label{app:Main_Experiment}
\subsubsection{Model and Training Configuration}

All experiments are conducted on top of the Qwen2.5 family of pretrained causal
language models. Unless otherwise specified, we adopt the 14B parameter variant
with bfloat16 precision.

\paragraph{Consistency Weight $\lambda$.} 
To balance the task-specific empirical risk $\mathcal{L}_{\mathrm{task}}$ and the consistency regularizer $\mathcal{L}_{\mathrm{cons}}$ within the total objective $\mathcal{L}_{\mathrm{total}} = \mathcal{L}_{\mathrm{task}} + \lambda \, \mathcal{L}_{\mathrm{cons}}$, we set $\lambda = 0.05$ across our main experiments. 

\paragraph{LoRA Configuration.}
CD-LoRA applies multiple low-rank adapters to the projection layers
(\textbf{gate\_proj}, \textbf{down\_proj}, and \textbf{up\_proj}) of the feed-forward
network. We use a LoRA rank of $r=4$, scaling factor $\alpha=32$, and a default
dropout rate of $0.1$ (or $0.2$ in selected ablations). The number of parallel
LoRA branches is set to $K=3$.

\paragraph{Routing and Consistency.}
When enabled, a lightweight router predicts per-token mixture weights over
the $K$ LoRA branches using a softmax-normalized linear projection. CD-LoRA
introduces a cross-projection consistency regularization that aligns the latent
representations produced by different LoRA branches during training.

\paragraph{Optimization Details.}
We train for one epoch over the combined multi-task dataset using the AdamW
optimizer with a peak learning rate of $2\times10^{-4}$ and a cosine learning
rate schedule. The warmup ratio is set to $0.03$, and weight decay is disabled.
To accommodate large models, we use gradient accumulation with 64 steps and
a per-device batch size of 3.

\paragraph{R-Drop Regularization}

To further stabilize training, we optionally incorporate R-Drop regularization.
During training, the LoRA-A projection is evaluated twice with independent
dropout masks, and a symmetric divergence term is computed between the resulting
latent representations. Unlike standard R-Drop formulations that operate on
output probabilities, we implement the regularization directly in the adapter
latent space, using an analytic KL divergence under a diagonal Gaussian
assumption. The resulting penalty is scaled by a coefficient $\alpha_{\text{R-Drop}}$
and added to the standard language modeling loss.

\paragraph{Data Processing and Sampling}

All datasets are pre-tokenized using the corresponding Qwen2.5 tokenizer with
a maximum sequence length of 512. For computational efficiency, we optionally
subsample each task to a fixed number of examples (e.g., 8,000 per task in the
12-task setting). When enabled, datasets are concatenated and randomly shuffled
before training to avoid task-order bias.

\paragraph{Reproducibility}

We fix random seeds for data sampling, model initialization, and training
($\texttt{seed}=7$ unless otherwise stated). All hyperparameters not explicitly
varied in ablation studies are held constant across experiments. Checkpoints are
saved every 1,500 steps, and evaluation is performed using the same decoding and
metric configurations reported in the main paper.
\subsection{BBH Evaluation with 12-Task Training}
\label{app:BBH_setup}
\subsubsection{Experiment Settings}

All models are based on Qwen2.5-7B and trained on the same 12-task unified benchmark~\ref{app:Twelve-Task} as in the main experiments (Section~\ref{app:Main_Experiment}). The evaluation of BBH~\ref{app:BBH} follows identical hyperparameters, optimization settings, and hardware configurations to ensure a fair comparison. All other experimental details—such as learning rate, batch size, LoRA rank, and dropout—are kept consistent with those in the main experiment (Section~\ref{app:Main_Experiment}).

\subsubsection{Experiment Results}

Table~\ref{tab:bbh_comparison_final} reports BBH performance under the four variants from the ablation study, now augmented with explicit indicators for routing and consistency alignment.

\begin{table}[htbp]
\centering
\caption{
Performance comparison on the BBH benchmark with 12-task training.}
\setlength{\tabcolsep}{8pt}
\resizebox{\columnwidth}{!}{
\begin{tabular}{lcc ccc}
\toprule
\textbf{Method} 
& \textbf{Router} 
& \textbf{Consistency} 
& \textbf{BBH (Avg)} 
& \textbf{BBH (Ov.)} 
& \textbf{Mem (MB)} \\
\midrule
R-LoRA\textsuperscript{$\dagger$}   
& \cmark & \xmark 
& 51.70 & 50.77 & 66070 \\

CR-LoRA\textsuperscript{$\ddagger$}  
& \cmark & \cmark 
& 52.50 & 51.69 & 80942 \\

DR-LoRA\textsuperscript{$\ast$}      
& \xmark & \xmark 
& 52.07 & 51.27 & 45510 \\

\textbf{CD-LoRA}\textsuperscript{$\S$} 
& \xmark & \cmark 
& \textbf{53.31} & \textbf{52.58} & 60288 \\
\bottomrule
\end{tabular}
}
\vspace{2mm}
\label{tab:bbh_comparison_final}
\end{table}

\section{Similarity Analysis on 14B Models}
\begin{figure}[t]
    \centering
    \includegraphics[width=0.5\linewidth]{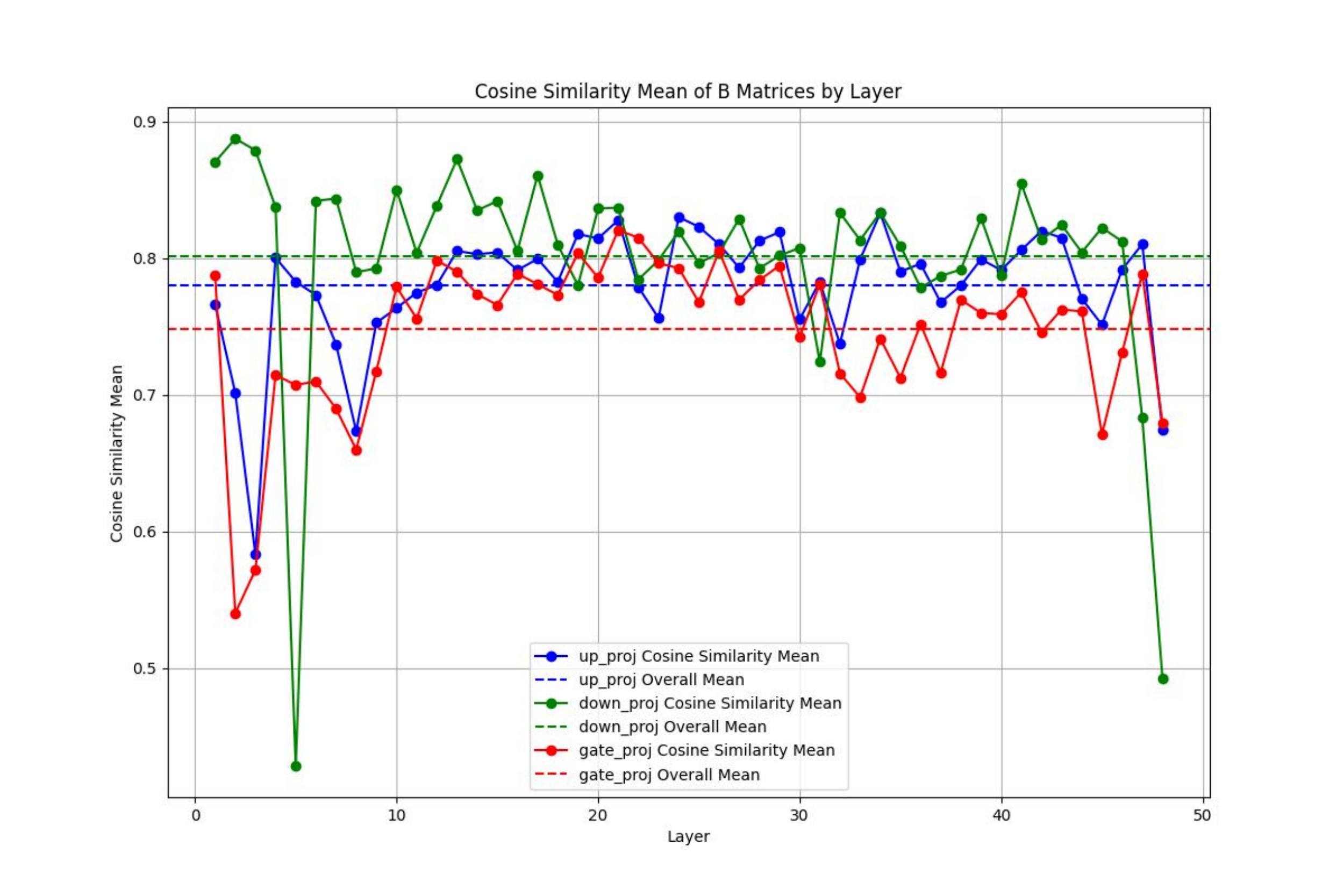}
    \caption{Layer-wise $B_i$ similarity matrix for R-LoRA on Qwen2.5-14B, showing fragmented task-specific subspaces induced by nonlinear routing.}
    \label{fig:R-LoRA_sim}
\end{figure}

\begin{figure}[t]
    \centering
    \includegraphics[width=0.5\linewidth]{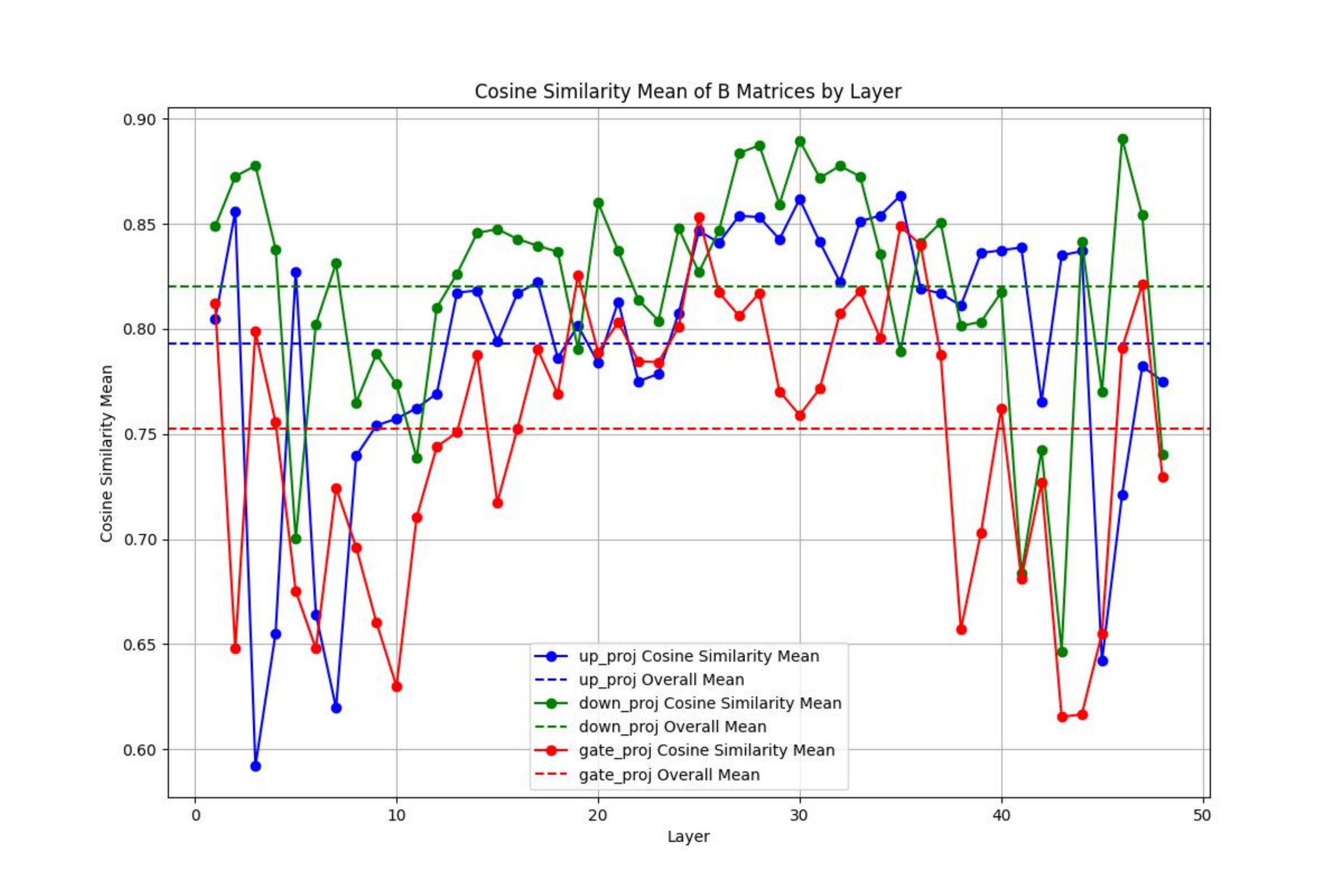}
    \caption{Layer-wise $B_i$ similarity matrix for CR-LoRA. While consistency is enforced, routing still leads to uneven and unstable subspace alignment.}
    \label{fig:CR-LoRA_sim}
\end{figure}


\end{document}